\documentclass{article}

\usepackage[preprint,nonatbib]{neurips_2024}

\usepackage[utf8]{inputenc} % allow utf-8 input
\usepackage[T1]{fontenc}    % use 8-bit T1 fonts
\usepackage{hyperref}       % hyperlinks
\usepackage{url}            % simple URL typesetting
\usepackage{booktabs}       % professional-quality tables
\usepackage{amsfonts}       % blackboard math symbols
\usepackage{nicefrac}       % compact symbols for 1/2, etc.
\usepackage{microtype}      % microtypography
\usepackage{xcolor}         % colors
\usepackage{amsmath}
\usepackage{bbm}

\usepackage{multirow}
\usepackage{ctable}
\usepackage{comment}
\usepackage{makecell}
\usepackage{tabularx}

\usepackage{bbding}
\usepackage{pifont}
\usepackage{wasysym}
\usepackage{amssymb}
\usepackage[flushleft]{threeparttable}

\newcommand{\cmark}{\ding{51}}%
\newcommand{\xmark}{\ding{55}}%

\title{A Classifier-Free Incremental Learning Framework for Scalable Medical Image Segmentation}

\author{
Xiaoyang Chen$^{1,2}$ \quad Hao Zheng$^{2}$ \quad Yifang Xie$^{3}$ \quad Yuncong Ma$^{2}$ \quad Tengfei Li$^{3}$\\
$^1$Duke University  \quad $^2$University of Pennsylvania \quad $^3$UNC Chapel Hill \\
\texttt{xiaoyang.chen@duke.edu} \\
}

\begin{document}

\maketitle

\begin{abstract}
Current methods for developing foundation models in medical image segmentation rely on two primary assumptions: a fixed set of classes and the immediate availability of a substantial and diverse training dataset. However, these can be impractical due to the evolving nature of imaging technology and patient demographics, as well as labor-intensive data curation, limiting their practical applicability and scalability. To address these challenges, we introduce a novel segmentation paradigm enabling the segmentation of a variable number of classes within a single classifier-free network, featuring an architecture independent of class number. This network is trained using contrastive learning and produces discriminative feature representations that facilitate straightforward interpretation. Additionally, we integrate this strategy into a knowledge distillation-based incremental learning framework, facilitating the gradual assimilation of new information from non-stationary data streams while avoiding catastrophic forgetting. Our approach provides a unified solution for tackling both class- and domain-incremental learning scenarios. We demonstrate the flexibility of our method in handling varying class numbers within a unified network and its capacity for incremental learning. Experimental results on an incompletely annotated, multi-modal, multi-source dataset for medical image segmentation underscore its superiority over state-of-the-art alternative approaches.

\end{abstract}

\section{Introduction}
Medical image segmentation is crucial for disease diagnosis, treatment planning, monitoring and follow-up, creating personalized anatomical models, and biomedical research. The need to accurately delineate anatomical structures and pathological regions has driven the development of many segmentation approaches. However, existing methods are typically tailored for specific applications, using datasets that lack diversity and are annotated solely for the relevant structures. Consequently, the trained models often exhibit limited applicability, robustness, and generalizability. These drawbacks have led to a surge in the popularity of developing foundation models, designed to be highly versatile and generalizable, for medical image segmentation.

Two paradigms dominate the development of foundation models for medical image segmentation. The first paradigm is prompt-driven interactive segmentation, popularized by the Segment Anything Model (SAM) \cite{kirillov2023segment}. Recent advances in medical image segmentation have led to the development of various SAM variants, such as MedSAM \cite{ma2024segment} and SAM-Med2D \cite{cheng2023sam}. These variants offer remarkable flexibility and scalability by enabling interactive user input prompts, such as points and bounding boxes, for segmenting diverse objects within medical images. However, their offline training approach relies on the availability of large, diverse datasets, which can be challenging to acquire and may not adequately represent all possible use cases or scenarios. Moreover, the need for human intervention makes these methods less suitable for applications requiring high-throughput and automated segmentation. The second paradigm focuses on learning from incompletely annotated data for automated segmentation \cite{fang2020multi,shi2021marginal,zhang2021dodnet,liu2023clip,chen2023versatile}. Due to the labor-intensive nature of data annotation, curating a substantial, fully annotated dataset for training segmentation models, especially for foundation models that are designed to be versatile and inherently involve many classes, is costly. Therefore, this research direction explores the effective use of partial \cite{fang2020multi,shi2021marginal,zhang2021dodnet,liu2023clip} or even sparse labels \cite{chen2023versatile}, which are comparatively less expensive, significantly reducing annotation costs and facilitating the development of foundation models. Unlike interactive segmentation, these methods automatically and consistently generate segmentation results without human involvement. However, they are also designed for offline training, and their typically fixed network architecture, determined by the total number of classes, limits their ability to handle an increasing number of classes.

In this study, we aim to devise a scalable approach for training foundation models in automatic medical image segmentation, concurrently addressing multiple factors that hinder scalability. Firstly, a large and diverse dataset is essential for training foundation models, but obtaining full annotations is costly and difficult to scale. Secondly, the network architecture of conventional methods depends on the total number of classes and lacks the flexibility to accommodate a varying number of classes, especially as the number of classes increases. Thirdly, offline training is ill-suited for training foundation models. Just as Rome wasn't built in a day, constructing a comprehensive dataset for training foundation models is a gradual process. Retraining the model offline from scratch whenever new data is added leads to inefficiencies and increased computational costs. Moreover, due to memory constraints and data privacy concerns, retaining previous training data when updating the model can be impractical. Consequently, the model should be able to adapt to the dynamic nature of data collection by continuously learning without relying on access to previous data.

To mitigate annotation costs and fully leverage available data, we base our approach on \cite{chen2023versatile}, utilizing model self-disambiguation capabilities and requiring only partially and/or sparsely labeled data for training. These data types are considerably more cost-effective and accessible compared to fully annotated data. To address network architecture limitations, we introduce a novel segmentation paradigm enabling the segmentation of a variable number of classes within a single classifier-free network, designed independently of class number. Furthermore, to enable gradual adaptation to new data and continuous capability expansion without the need to retain previous data, we integrate our classifier-free network with a knowledge distillation-based incremental learning framework. This integration facilitates the gradual assimilation of new information from non-stationary data streams while mitigating the risk of catastrophic forgetting. In summary, our contributions are as follows:
\begin{itemize}
\item We propose a scalable approach for training foundation models in automatic medical image segmentation using partially and/or sparsely labeled data. %Scalability challenges associated with data availability, network architecture, and no-replay incremental learning are comprehensively addressed.
\item We present a novel segmentation paradigm that enables flexible segmentation of a variable number of classes within a single classifier-free network.
\item We integrate the classifier-free network with knowledge distillation-based incremental learning to adapt gradually to new data and prevent catastrophic forgetting.
\item We show our method's superior scalability and reduced risk of catastrophic forgetting compared to existing methods, leveraging a multi-source, multi-modality dataset.
\end{itemize}

\section{Related Works}
\subsection{Paradigms of Foundation Models in Medical Image Segmentation}
\noindent \textbf{Prompt-Driven Interactive Segmentation} Traditional interactive segmentation methods leverage user-provided hints, such as clicks \cite{liu2023simpleclick}, scribbles \cite{lin2016scribblesup,wong2023scribbleprompt}, or bounding boxes \cite{rajchl2016deepcut}, to iteratively and accurately delineate the structures of interest within an image. These methods are typically trained offline on datasets with a limited number of classes and images and are specialized. Recently, SAM \cite{kirillov2023segment}, trained offline on a massive dataset, was introduced as a general-purpose foundation model for image segmentation tasks, requiring minimal or no user input. While SAM can handle a wide variety of tasks out of the box, its precision might not match that of interactive methods in highly specialized tasks, such as medical imaging. To address this issue, MedSAM \cite{ma2024segment} and SAM-Med2D \cite{cheng2023sam}, fine-tuned on datasets specifically for medical image segmentation, have emerged.

While the class-agnostic feature of SAM and its variants offers flexibility by not requiring annotation for all structures of interest in training images, each annotated structure must still be fully delineated. However, this strict requirement drives up the cost of obtaining training data, limiting the model's adaptability to new use cases or scenarios. Moreover, the necessity for human intervention to segment each instance and/or assign labels to the segments makes this method less ideal for applications demanding high-throughput, automated segmentation. %Fine-tuning the model on a different dataset poses a risk of losing previously acquired knowledge, thereby compromising its efficacy as a foundation model.

\noindent \textbf{Segmentation with Incompletely Annotated Data} Weakly supervised segmentation \cite{fang2020multi,shi2021marginal,chen2023versatile} and dynamic models \cite{zhang2021dodnet,liu2023clip} leverage incompletely annotated data for training, offering promising avenues for building foundation models for automated medical image segmentation. However, semantic ambiguity within the labels presents challenges. Dynamic models \cite{zhang2021dodnet,liu2023clip} handle partial labelling ambiguity by addressing one segmentation task at a time, adjusting the network output via a controller. However, these models face optimization challenges and exhibit low inference efficiency \cite{chen2023versatile}. Weakly supervised learning methods in \cite{fang2020multi} and \cite{shi2021marginal} blend a small quantity of fully annotated data with a larger quantity of partially labeled data to train a unified multi-class segmentation model. However, scalability is hampered by the requirement for fully annotated data when incorporating new classes. Chen et al. \cite{chen2023versatile} tackle this constraint by proposing a method that empowers deep neural networks to self-disambiguate even when trained without any fully annotated data.

While promising for training foundation models, current methods in this line assume a static dataset and undergo offline training, overlooking the dynamic nature of real-world data. Additionally, their network architecture is usually closely tied to prediction classes, making them inflexible in accommodating an expanding class set.

\subsection{Incremental Learning}

\noindent \textbf{Dynamic Network Expansion} Dynamic network expansion methods dynamically adapt the model architecture as new tasks are added. Progressive neural networks \cite{rusu2016progressive} add a new network column for each task, keeping previously learned parameters fixed and allowing knowledge transfer via lateral connections. Dynamically expandable networks \cite{yoon2017lifelong} expand the model by selectively adding neurons and connections based on the new tasks' requirements, while pruning unnecessary parts to maintain efficiency. However, the increasing number of parameters and memory consumption as more tasks are added limit scalability. Additionally, they require knowledge of the task identity to select the appropriate column during inference, which may not be available in many practical scenarios.

\noindent \textbf{Rehearsal and Pseudo-Rehearsal} Rehearsal methods \cite{rebuffi2017icarl} typically store a small subset of training data from previous tasks and integrate it with the current task data to optimize network parameters. However, they face potential limitations due to memory constraints and concerns about data privacy and security. In contrast, pseudo-rehearsal methods \cite{shin2017continual} train a generative model to produce synthetic data that resembles the data from previously learned tasks, pairing it with pseudo-labels derived from the old model to train the new one. While this approach has been successful on image classification tasks with toy datasets like MNIST, it has yet to be demonstrated for medical image segmentation. %In contrast, pseudo-rehearsal methods \cite{shin2017continual} train a generative model to generate synthetic data resembling the data from previously learned tasks. The synthetic data is paired with pseudo-labels derived from the old model to train the new one. While successful on image classification toy datasets like MNIST, this approach has yet to be demonstrated for medical image segmentation tasks. %the method here overcomes catastrophic forgetting by using pseudo image-label pairs vs. my method demonstrates that catastrophic forgetting can be obviated by simply using unlabeled data)

\noindent \textbf{Parameter Regularization} Regularization-based methods add terms to the loss function during training to penalize changes to parameters that are important for previously learned tasks. Elastic weight consolidation \cite{kirkpatrick2017overcoming} uses the Fisher information matrix to measure the importance of each parameter. Memory aware synapses \cite{aljundi2018memory} calculate weights by accumulating the sensitivity of the predicted output function to changes in each parameter. However, penalizing adjustments to important parameters may impede the network's capability in learning new tasks. Parameters previously deemed unimportant might become crucial in acquiring new knowledge, and the change in their values may negatively affect the retention of old information.

\noindent \textbf{Knowledge Distillation} Knowledge distillation-based incremental learning methods \cite{li2017learning,cermelli2020modeling,liu2021incremental} involve transferring knowledge from a previously trained model (i.e., the teacher model) to a new model (i.e., the student model) being trained on new tasks. Regularization terms are incorporated to encourage the student model's predictions on old classes to match the soft labels generated by the teacher model. However, studies have demonstrated that knowledge distillation alone may not effectively prevent catastrophic forgetting, particularly when the training data for the student and teacher models originate from different distributions \cite{aljundi2017expert}. To address this limitation, a combination of knowledge distillation and rehearsal is often employed \cite{rebuffi2017icarl}. In contrast to this hybrid approach \cite{rebuffi2017icarl}, our approach aims to retain previous knowledge without storing any old data, even during continual training on data from disparate domains with no overlap with previous classes, achieving enhanced scalability and generalizability. %We identify the pitfalls that affect knowledge preservation.

%These methods have been shown to be effective and advantageous in continual learning of medical images. Compared with rehearsal-based methods, distillation-based methods require no additional storage and avoid privacy invasion. Compared with isolation-based methods, they have a higher upper limit of learning ability because they do not limit the number of active parameters in the network to train the new data. Compared with regularization-based methods, they guide the learning of the new model according to the knowledge of the old model without reducing its learning ability on the new data.

%Drawbacks of conventional approaches for image classification: 1) Network outputs are tightly coupled with prediction classes. When new classes are added, structural changes to the architecture are required, i.e., new neurons added to accommodate for the new classes. 2) In order to acquire a final prediction from a multi-head network, outputs need to be aggregated. 3) The updated model is expected to be biased in the predictions that favor new classes. [from paper Semantic Drift Compensation for Class-Incremental Learning]

\section{Methods}\label{methods}

\begin{figure}[t]
  \centering
  \includegraphics[width=0.9\textwidth]{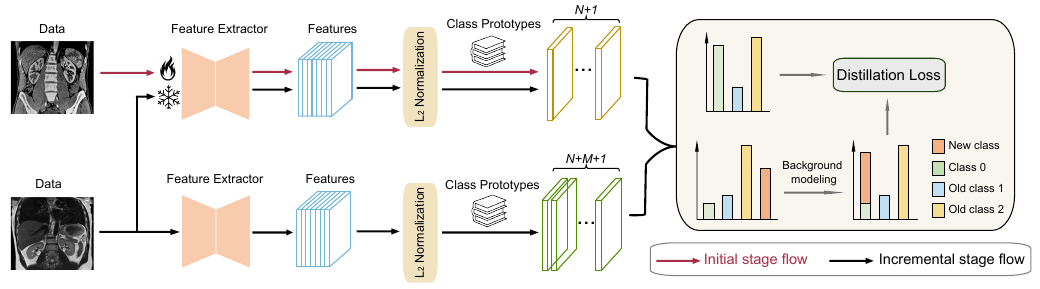}
  \caption{Overview of our approach. It incorporates a classifier-free, teacher-student architecture for incrementally learn from data streams. Only the teacher model is involved and trained in the initial stage. In the incremental stage, both models are involved, but the teacher model is frozen.}
  \label{fig1}
\end{figure}

\subsection{Classifier-Free Architecture}
Neural networks for image segmentation typically include a feature extractor and a classifier. The classifier, located in the final layers of the network, typically consists of one or more convolutional layers with isotropic kernel sizes of $1$, followed by an appropriate activation layer, such as a softmax layer for multi-class segmentation. The number of output channels in the final convolutional layer corresponds to the total number of classes to be segmented. Consequently, incorporating new classes necessitates structural alterations to the architecture. Although widely adopted and successful, this design approach proves inflexible in adapting to a growing class set within the same model. Recognizing this inherent limitation, we propose a classifier-free architecture, retaining only the feature extractor. This eliminates the reliance on a preset number of target classes.

To train the classifier-free image segmentation model, we propose using a predefined set of class prototypes. Each prototype is a unit vector with the same length as the output feature representations generated by the feature extractor. Additionally, we employ a contrastive representation learning approach, as outlined below, to facilitate model training. This approach is designed to ensure that the feature representations for each pixel/voxel align with their respective class prototypes.

\subsection{Class Prototypes}
While it is possible to choose class prototypes as random noises, we embrace a more principled approach: leveraging a pretrained CLIP model \cite{radford2021learning} with the ViT-B/32 architecture. We tokenize text representing each class and pass it through the CLIP model to generate an embedding vector of length 512. For example, to derive the liver prototype, we input the text ``liver'' into the CLIP model. %The resulting embedding vector is then $L_{2}$-normalized to obtain the liver category prototype.

Let $\mathbf{v}_i$ be the CLIP embedding for class $i$ (note that class $0$ is a special class that includes the real background and any future classes not yet involved in training) and let $\mathbf{f}_i$ be the feature representation for the $i$-th pixel/voxel. The length of $\mathbf{f}_i$ and $\mathbf{v}_i$ should be equal. To manage memory consumption, we employ a variance-based feature selection method to extract a subset of $64$ features from $\{ \mathbf{v}_i \}_{i=0}^{N}$, where $N$ is the number of structures of interest,  retaining only those with variance exceeding the specified threshold. We observe that model training and feature representation interpretability benefit from vector orthogonalization. Therefore, we used the Gram-Schmidt algorithm to obtain a set of unit-length orthogonal vectors $\{ \mathbf{e}_i \}_{i=0}^{N}$ from $\{ \mathbf{v}_i \}_{i=0}^{N}$, as the final class prototypes.

\subsection{Contrastive Representation Learning}\label{loss_section}
The training objective is to align $\mathbf{f}_i$ (after $L_2$ normalization) for the $i$-th pixel/voxel with the corresponding class prototype $\mathbf{e}_{y_i}$, where $y_i$ is the label for the pixel/voxel. The similarity between $\mathbf{f}_i$ and any class prototype $\mathbf{e}_j$ is measured by their cosine similarity, $\langle \mathbf{f}_i, \mathbf{e}_j \rangle$. We train our model by converting similarity scores, scaled by a temperature parameter $\tau$ to control the concentration level of the distribution, into numbers that we interpret as probabilities using the following function:
\begin{equation}
p_{ij} = \frac{e^{\langle \mathbf{f}_i, \mathbf{e}_j \rangle / \tau}}{ \sum_{k=0}^{N} e^{\langle \mathbf{f}_i, \mathbf{e}_k \rangle / \tau} },
\end{equation}
where $p_{ij}$ represents the probability of $i$-th pixel/voxel belonging to class $j$. The rationale is that each feature representation $\mathbf{f}_i$ is compared to each class prototype to obtain a similarity score, with higher probabilities assigned to more similar pairs and lower probabilities to less similar pairs.

Note that the class number $N$ is involved in loss computation but not in the network architecture, making our network architecture independent of the class number. The value of $N$ can be adjusted for tasks involving different numbers of classes.

With $p_{ij}$, we apply the ambiguity-aware focal cross-entropy loss ($\mathcal{L}_{\text{focal\_ce}}$) and dice loss ($\mathcal{L}_{\text{dice}}$) as proposed in \cite{chen2023versatile} for training with partially and/or sparsely labeled data. Partially labeled data have a subset of structures annotated in the entire volume, while sparsely labeled data have only a fraction of their slices annotated, with each slice containing annotations for only a subset of structures. Generally, this subset can comprise any combination and be denoted as $\Phi_{M}$ $=$ $\{i_{1}, i_{2}, \dots, i_{M}\}$, where $1 \leq i_{p} < i_{q} \leq N$ for any $p < q$. The loss for partially labeled data is calculated volumetrically, while for sparsely labeled data, it is computed on a slice-by-slice basis.
\begin{equation}
\mathcal{L}_{\text{focal\_ce}} = \frac{1}{N_{v}} \sum_{c \in \{0\} \cup \Phi_{M}}\sum_{i=1}^{N_{v}} \mathbbm{1}_{y_{i}=c} (1-\tilde{p}_{ic})^2\log \tilde{p}_{ic},
\end{equation}
\begin{equation}
\mathcal{L}_{\text{dice}}=1-\frac{1}{|\Phi_{M}|+1}  \sum_{c \in \{0\} \cup \Phi_{M}} \frac{2 \cdot \sum_{i=1}^{N_{v}} \tilde{p}_{ic} \tilde{y}_{ic} + \epsilon}{2 \cdot \sum_{i=1}^{N_{v}} \tilde{p}_{ic} \tilde{y}_{ic} + \sum_{i=1}^{N_{v}} \tilde{p}_{ic} (1-\tilde{y}_{ic}) + \sum_{i=1}^{N_{v}} (1-\tilde{p}_{ic}) \tilde{y}_{ic} + \epsilon},
\end{equation}
where $\mathbbm{1}$ denotes an indicator function, $|\cdot|$ is the cardinality, $N_{v}$ represents the number of pixels in the slice for sparsely labeled data or the number of voxels in the volume for partially labeled data, $\epsilon$ is set to $1$ to avoid division by 0,
\begin{equation}
\begin{aligned}
\tilde{p}_{ic}=
\begin{cases}
p_{ic}, &c \in \Phi_{M},\\
\sum_{j \not\in \Phi_{M}} p_{ij}, &c = 0,
\end{cases}
\quad
\tilde{y}_{ic}=
\begin{cases}
y_{ic}, &c \in \Phi_{M},\\
\sum_{j \not\in \Phi_{M}} y_{ij}, &c = 0,
\end{cases}
\end{aligned}
\end{equation}
$p_{ic}$ and $y_{ic}$ represent the $c$-th element of the probability vector and the one-hot encoded vector for the expert label for the $i$-th pixel, respectively.

\subsection{Regularizations}\label{reg_section}
\noindent \textbf{Entropy Minimization} Following \cite{chen2023versatile}, we minimize the Shannon entropy during training, encouraging the model to produce more confident and informative predictions.
\begin{equation}
\mathcal{L}_{\text{entropy}} = -\frac{1}{N_{v}} \sum_{i=1}^{N_{v}} \sum_{c=0}^{N} p_{ic} \log p_{ic},
\end{equation}
\noindent \textbf{Volume Minimization} We observed that the model may struggle to self-disambiguate when trained solely with 
$\mathcal{L}_{\text{focal\_ce}}$, $\mathcal{L}_{\text{dice}}$ and $\mathcal{L}_{\text{entropy}}$ if the proportion of annotated classes in a slice or volume is extremely small. Empirically, we found that adding the following volume regularization term, which encourages the volume of structures of interest (excluding class $0$) to be small during early stages of training, helps the model escape local minima.
\begin{equation}
\mathcal{L}_{\text{volume}} = -\frac{1}{N} \sum_{c=1}^{N} \sum_{i=1}^{N_{v}}  p_{ic},
\end{equation}

\subsection{Knowledge Distillation-based Incremental Learning}
In the context of incremental learning, training occurs across multiple phases or steps, with each step introducing new categories to be learned. Consider a scenario where a pre-existing model, referred to as the teacher model, has been trained on a dataset with label set $\Phi_{N}=\{j \mid 1 \leq j \leq N\}$ using the loss functions and regularization terms present in subsections \ref{loss_section} and \ref{reg_section} , and a new model, referred to as the student model, is tasked with training on data with a new set of labels $\Phi_{N+1}^{N+M}=\{j \mid N+1 \leq j \leq N+M\}$, which includes $M$ new classes that are not present in $\Phi_{N}$. The student model is expected to not only segment the new classes in $\Phi_{N+1}^{N+M}$ but also the classes in $\Phi_{N}$. The teacher model acts as a source of knowledge on classes in $\Phi_{N}$ to the student model. Note that while we assume disjoint label sets for the previous and current training data, this scenario is not limiting. If the training data for the student model include annotations for classes in $\Phi_{N}$, we can exclude those voxels when computing the knowledge distillation loss ($\mathcal{L}_{\text{KD}}$) below.
\begin{equation}
\mathcal{L}_{\text{KD}} = \frac{1}{N_{v}} \sum_{i=1}^{N_{v}} \sum_{c=0}^{N} p_{ic}^{t} \log \frac{p_{ic}^{t}}{\hat{p}_{ic}^{s}},
\end{equation}
where
\begin{equation}
\hat{p}_{ic}^{s}=
\begin{cases}
p_{ic}^{s}, &c \in \Phi_{N},\\
\sum_{j \in \{0\} \cup \Phi_{N+1}^{N+M}} p_{ij}^{s}, &c = 0,
\end{cases}
\end{equation}
$p_{ij}^{t}$ and $p_{ij}^{s}$ represent the probability of the $i$-th pixel/voxel belonging to class $j$, as derived from the teacher and student networks, respectively. The superscripts ``t'' and ``s'' denote teacher and student. Note that the temperature parameter $\tau$ can differ between the teacher and student models.

\subsection{Objective Function}
The objective function ($\mathcal{L}$) is a weighted sum of $\mathcal{L}_{\text{focal\_ce}}$, $\mathcal{L}_{\text{dice}}$, $\mathcal{L}_{\text{reg}}$, $\mathcal{L}_{\text{volume}}$ and $\mathcal{L}_{\text{KD}}$:
\begin{equation}
\mathcal{L} = \mathcal{L}_{\text{focal\_ce}} + \mathcal{L}_{\text{dice}} + \lambda_{1}\mathcal{L}_{\text{entropy}} + \lambda_{2}\mathcal{L}_{\text{volume}} + \lambda_{3}\mathcal{L}_{\text{KD}}.
\end{equation}
$\lambda_{1}$ is set to $3$ for inputs without annotations to mitigate the null effects of $\mathcal{L}_{\text{focal\_ce}}$ and $\mathcal{L}_{\text{dice}}$, and $1$ otherwise. $\lambda_{2}$ is defined as $10^{-5} \times \text{max}\{ \frac{5 - \text{epoch}}{5}, 0\}$, where $epoch$ is epoch number, and one epoch includes $1000$ iterations. $\lambda_{3}$ equals $0$ in initial stage and $1$ during incremental stages. Note that the computation of $\mathcal{L}_{\text{focal\_ce}}$ and $\mathcal{L}_{\text{dice}}$ differs between the initial stage and incremental stages. Assuming disjoint label sets, $\Phi_{M}$ should be replaced with $\Phi_{N+1}^{N+M}$ in Equations 2 and 3 in incremental stages.

\section{Experiments and Results}

\begin{table}[h]
        \centering
        \caption{Information about the number of training, testing, and unlabeled images, as well as the labeled structures, across different datasets.}
        \resizebox{0.66\columnwidth}{!}{%
        \begin{tabular}{c|c|c|c|c}
        \toprule
        \textbf{Dataset} & \textbf{\# Training} & \textbf{\# Testing} & \textbf{\# Unlabeled} & \textbf{\# Labeled Structures} \\
        \midrule
        AbdomenCT-1K  & 893 & 100 & 0 & 5 \\
        AMOS-CT & 270 & 30 & 199 & 15 \\
        AMOS-MRI & 54 & 6 & 41 & 15 \\
        BTCV & 27 & 3 & 0 & 13 \\
        FLARE22 & 45 & 5 & 0 & 13 \\
        NIH-Pancreas & 73 & 9 & 0 & 1\\
        TotalSegmentator & 1081 & 122 & 0 & 29\\
        Urogram & 109 & 13 & 0 & 3\\
        WORD & 100 & 20 & 0 & 12\\
        LifespanBTS & 67 & 8 & 0 & 3\\
        \bottomrule
    \end{tabular}}
    \label{tab1}
\end{table}

\subsection{Experimental Setup}

\noindent \textbf{Datasets and Preprocessing} We curated a dataset of 3,035 volumetric images from various sources: AbdomenCT-1K \cite{ma2021abdomenct}, AMOS \cite{ji2022amos}, BTCV \cite{landman2015miccai}, FLARE 2022 (FLARE22) \cite{ma2023unleashing}, NIH pancreas \cite{roth2015deeporgan}, TotalSegmentator \cite{wasserthal2022totalsegmentator}, WORD \cite{luo2021word}, a private CT urogram dataset (Urogram), and a dataset for lifespan brain tissue segmentation (LifespanBTS) \cite{chen2023brain}. The AMOS dataset is divided into AMOS-CT and AMOS-MRI subsets based on image modalities. Our dataset includes $33$ anatomical structures: spleen, right kidney (RK), left kidney (LK), gallbladder (GB), esophagus (Eso), liver, stomach (St), aorta, postcava (PC), pancreas (Pan), right adrenal gland (RAG), left adrenal gland (LAG), duodenum (Duo), bladder, prostate/uterus (PU), portal vein and splenic vein (PSV), left clavicle (LC), right clavicle (RC), colon, left femur (LF), right femur (RF), heart, left hip (LH), right hip (RH), left humerus (LHU), right humerus (RHU), lung, rib, trachea, vertebrae, gray matter (GM), white matter (WM), and cerebrospinal fluid (CSF). We retained only the masks for the structures of interest from TotalSegmentator and WORD. Additionally, we semi-automatically divided the masks for kidneys and adrenal glands into left and right sides in WORD and AbdomenCT-1K. For cases with only one kidney or adrenal gland, we manually assigned the correct labels. We have removed images that are corrupted or have incorrect headers. Images were randomly split into training and testing data.

We followed the naming conventions and used the recommended hierarchical sampling approach proposed in \cite{chen2023versatile} for training data generation. Additionally, we standardized all images to a common coordinate system to facilitate model training with images in varied orientations. All data were resampled to a uniform spacing of $2\times2\times2$ $\text{mm}^{3}$. Intensity values in CT images were clipped at $-400$ and $400$ HU, while MRI images were clipped at the 1st and 99th percentiles of their intensity distributions. Finally, the intensity values were normalized to the range of $[0, 1]$.

\noindent \textbf{Implementation details} Implementation was done using PyTorch. We utilized the AdamW optimizer \cite{loshchilov2017decoupled} with an initial learning rate of $0.001$ and a polynomial learning rate scheduler with a decay rate of $0.9$. During training, data augmentations such as random rotation and scaling were applied. Unless stated otherwise, the default patch size was set to $112\times112\times112$ and the number of iterations to $200,000$. Distributed data parallelism was employed to improve training efficiency, conducted on a single node equipped with 8 NVIDIA Titan Xp GPUs. The effective batch size was $8$ for all experiments, and all models were trained from scratch.

\noindent \textbf{Evaluation metrics} Dice Similarity Coefficient (DSC, $\%$) was used as the evaluation metric. Different methods and settings were compared using 1) per-subject mean DSC, which computes the average DSC for all annotated classes for each testing subject and then averages these values across all subjects, and 2) per-structure mean DSC, which involves averaging the DSCs for each specific structure across all individual images where that structure is annotated.

\noindent \textbf{Competing methods} We compared our method with DoDNet \cite{zhang2021dodnet}, the CLIP-driven model (CLIP-driven)\cite{liu2023clip}, and the versatile model (Versatile) \cite{chen2023versatile}, all of which effectively utilize incompletely labeled data for automatic medical image segmentation and demonstrate potential for universal segmentation. For fair comparisons, all methods used the same 3D TransUNet architecture as in \cite{chen2023versatile} and adopted the same hierarchical sampling approach for training. Each method has only minimal necessary differences in their architecture to accommodate their unique features.

\subsection{Main Results}
We conducted experiments to demonstrate: 1) the flexibility of the proposed classifier-free model in handling a varying number of classes within the same architecture, and 2) the efficacy and superiority of the proposed method in incremental learning settings.

\subsubsection{Offline Training on Varying Class Numbers}\label{offline_results}
To illustrate the flexibility of the proposed classifier-free model, we designed two segmentation tasks: a 16-class task and a 30-class task. The 16-class task involves segmenting the spleen, RK, LK, GB, Eso, liver, St, aorta, PC, Pan, RAG, LAG, Duo, bladder, PU, and PSV. The 30-class task extends this by adding 14 extra structures, including the LC, RC, colon, LF, RF, heart, LH, RH, LHU, RHU, lung, rib, trachea, and vertebrae. Both tasks utilize the same $8$ datasets for training and testing, including AbdomenCT-1K, AMOS (excluding unlabeled images), BTCV, FLARE22, NIH pancreas, TotalSegmentator, WORD, and Urogram. %Their differences.

Because DoDNet and the CLIP-driven model can only handle partially labeled data, not sparsely labeled data, we compared our method solely with the Versatile model on sparsely labeled data. To showcase our method's ability to handle sparsely labeled data, which is less costly and more widely available, we conducted experiments where we retained one out of every five slices (i.e., $20\%$ annotations) from axial, sagittal, or coronal views in the training data for segmenting the aforementioned $30$ structures.

\begin{table}[t!]
        \centering
        \caption{Method performance comparison on partially labeled data.}
        \resizebox{0.75\columnwidth}{!}{%
        \begin{tabular}{c|c|c|c|c|c}
            \toprule
            \multirow{2}{*}{\centering \textbf{Method}}  & \multicolumn{2}{c|}{\textbf{16-class Task}} & \multicolumn{2}{c|}{\textbf{30-class Task}} & \multirow{2}{*}{\centering \textbf{Flexible?}} \\
            & \textbf{Per-subject} & \textbf{Per-structure} & \textbf{Per-subject} & \textbf{Per-structure} & \\
            \midrule
            DoDNet \cite{zhang2021dodnet} & $84.2 (15.3)$ & $80.3 (18.1)$ & $85.2 (15.3)$ & $84.3 (17.4)$ & No \\
            CLIP-driven \cite{liu2023clip} & $83.0 (15.7)$ & $79.2 (19.2)$ & $85.0 (15.3)$ & $82.8 (17.1)$ & Yes \\
            Versatile \cite{chen2023versatile} & $88.7 (7.0)$ & $85.7 (11.9)$ & $89.1 (5.8)$ & $88.2 (10.3)$ & No \\
            Ours & $88.6 (7.1)$ & $85.4 (12.4)$ & $89.5 (5.6)$ & $88.5 (10.0)$ & Yes \\
            \bottomrule
         \end{tabular}}
        \label{tab2}
\end{table}

\noindent \textbf{Results on Partially Labeled Data}
Table~\ref{tab2} summarizes the segmentation performance for DoDNet, CLIP-driven, Versatile, and the proposed method. Our experimental results reveal that DoDNet achieves a per-subject mean DSC of $84.2\%$ and a per-structure mean DSC of $80.3\%$ on the 16-class task. However, its reliance on a one-hot vector to encode task information limits its flexibility. In contrast, the CLIP-driven model uses a fixed-length vector to encode task information, making it more flexible, but its performance drops by $1.2\%$ and $0.9\%$ in terms of per-subject and per-structure averages, respectively. Our proposed method, on the other hand, achieves a per-subject mean DSC of $88.6\%$ and a per-structure mean DSC of $85.4\%$ on the same task, significantly outperforming DoDNet and CLIP-driven by at least $4.4\%$ on both evaluation metrics. Moreover, our method's classifier-free feature makes it more flexible than the Versatile model, which achieves comparable performance but is limited in model scalability by its dependence on the number of classes.

The experimental results for the 30-class task show a similar pattern. The difference is that the proposed method exhibits superior performance on both a per-subject and per-structure basis.

%This evaluation method computed the average DSC of all annotated classes for each testing subject and subsequently averaged them across subjects. During training DoDNet and CLIP-driven, we observed that they exhibited significantly slower convergence rates and thus doubled the training time compared to our proposed method.

\begin{table}[ht]
        \centering
        \caption{Method performance comparison on sparsely labeled data ($20\%$ annotations).}
        \resizebox{0.56\columnwidth}{!}{%
        \begin{tabular}{c|c|c|c|c}
            \toprule
            \multirow{2}{*}{\textbf{Method}} & \multirow{2}{*}{\textbf{View}} & \multicolumn{2}{c|}{\textbf{30-class Task}} & \multirow{2}{*}{\textbf{Flexible?}}\\
            & & \textbf{Per-subject} & \textbf{Per-structure} & \\
            \midrule
            \multirow{3}{*}{Versatile} & axial & $88.0(6.9)$ & $86.5(12.2)$ & \multirow{3}{*}{No}\\
                                                    & sagittal & $88.5(6.3)$ & $87.1(10.7)$ & \\
                                                    & coronal & $88.3(6.3)$ & $86.7(11.4)$ & \\
            \midrule
            \multirow{3}{*}{Ours} & axial & $88.4(6.3)$ & $87.3(11.1)$ & \multirow{3}{*}{Yes}\\
                                              & sagittal & $88.8(6.0)$ & $87.9(10.0)$ & \\
                                              & coronal & $88.8(6.1)$ & $87.7(10.4)$ & \\
            \bottomrule
         \end{tabular}}
        \label{tab3}
\end{table}

\noindent \textbf{Results on Sparsely Labeled Data}
The experimental results on sparsely labeled data are presented in Table~\ref{tab3}. Compared with the Versatile model, the proposed method consistently exhibits superior segmentation performance across views and evaluation metrics. The most notable improvement is observed when evaluated on a per-structure basis, where the improvement is at least $0.8\%$. Importantly, we observed that our models trained with only $20\%$ annotations can already outperform DoDNet and CLIP-driven models by a large margin, which were trained with all slices used (cf. Table~\ref{tab2}).

\subsubsection{Class- and Domain-Incremental Learning without Rehearsal}
The proposed classifier-free model eliminates the dependency on the number of classes in network architecture design, making it flexible enough to accommodate a varying number of classes. This flexibility makes it particularly suitable for class-incremental learning, where the label set expands with each incremental step. We designed two different incremental learning tasks: a class-incremental learning task (Task I) and a combined class- and domain-incremental learning task (Task II), where not only the number of classes increases but the data also comes from different distributions.

Note that for both tasks, the objective is for the student model to not only acquire new concepts but also preserve the knowledge learned by the teacher model, all without relying on any data (or more precisely, annotations) utilized in the initial stage. In our experiments on both tasks, the patch size was adjusted to $96\times96\times96$ due to limited GPU memory.

\noindent \textbf{Results on Task I} For Task I, we first trained an initial model to segment the same $16$ classes as described in the 16-class task in Section~\ref{offline_results}, using the same $8$ datasets. Then, we continually trained the model on an additional set of $14$ classes using solely TotalSegmentator, utilizing only the labels specific to these classes. That is, we deliberately reset the labels for the $16$ classes trained in the initial stage to $0$ within the training data.

The results for Task I are presented in Table~\ref{tab4}. It is evident that our proposed method exhibits the least susceptibility to catastrophic forgetting, attaining a per-structure mean DSC of $84.4\%$ after the incremental stage, which surpasses that of the other three compared methods. Moreover, it outperforms the second-best Versatile model by a notable margin of $2.6\%$. Additionally, we observed that the quantity of training images during the incremental stage is influential. As depicted in Table~\ref{tab5}, after the incremental stage, performance on classes $1$--$16$ consistently improves with an increase in the number of training images during the incremental stage. Unsurprisingly, performance on classes $17$--$30$ also improves as more training images are utilized in the incremental stage.

\begin{table}[t]
        \centering
        \caption{Performance across the initial stage, incremental stage, and upper bound for Task I.}
        \resizebox{0.9\columnwidth}{!}{%
        \begin{tabular}{c|c|c|c|c}
            \toprule
            \multirow{2}{*}{\centering \textbf{Method}}  & \textbf{Initial Stage} & \multicolumn{2}{c|}{\textbf{Incremental Stage}} & \textbf{Upper Bound} \\
            & \textbf{Per-structure (1--16)} & \textbf{Per-structure (1--16)} & \textbf{Per-structure (17--30)} & \textbf{Per-structure (17--30)} \\
            \midrule
            DoDNet \cite{zhang2021dodnet} & $79.3 (18.4)$ & $73.4 (16.9)$ & $86.9 (16.2)$ & $87.9 (16.7)$ \\
            CLIP-driven \cite{liu2023clip} & $80.3 (17.2)$ & $75.5 (16.7)$ & $83.7 (16.4)$ & $83.7 (16.4)$ \\
            Versatile \cite{chen2023versatile} & $85.1 (12.5)$ & $81.8 (13.0)$ & $91.7 (7.5)$ & $91.9 (7.3)$ \\
            Ours & $85.3 (12.3)$ & $84.4 (13.1)$ & $92.2 (6.9)$ & $92.2 (7.2)$ \\
            \bottomrule
         \end{tabular}}
        \label{tab4}
\end{table}

\begin{table}[ht]
        \centering
        \caption{Results with different portions of training data used for the incremental stage of Task I.}
        \resizebox{0.72\columnwidth}{!}{%
        \begin{tabular}{c|c|c|c}
            \toprule
            \multirow{2}{*}{\centering \textbf{\# Training Images}}  & \textbf{Initial Stage} & \multicolumn{2}{c}{\textbf{Incremental Stage}} \\
            & \textbf{Per-structure (1--16)} & \textbf{Per-structure (1--16)} & \textbf{Per-structure (17--30)} \\
            \midrule
            5\% & $\multirow{3}{*}{\centering 85.3 (12.3)}$ & $83.0(14.0)$ & $91.1(8.3)$ \\
            20\% & & $83.9(13.3)$ & $91.6(7.8)$ \\
            100\% & & $84.4 (13.1)$ & $92.2 (6.9)$ \\
            \bottomrule
         \end{tabular}}
        \label{tab5}
\end{table}

\noindent \textbf{Results on Task II} For Task II, we trained the initial stage similarly to Task I but used a different set of class prototypes. During the incremental stage, we trained the model using LifespanBTS, which contains $3$ additional classes. In our experiments, we discovered that catastrophic forgetting persisted despite the losses and regularizations described in Sec.~\ref{methods} when confronted with data from completely different distributions lacking any instances to the concepts learned by the initial model. As shown in Table~\ref{tab6}, the per-structure average DSC dropped from $85.2\%$ to $0\%$ when no unlabeled data from AMOS was used during the incremental stage, indicating that the new model lost all the knowledge gained by the teacher model. However, after we augmented our dataset with $240$ unlabeled images from AMOS, the student model was able to retain the initial stage knowledge effectively, achieving an average per-structure DSC of $81.6\%$. Note that these unlabeled images are distinct from the training and testing images in the AMOS dataset used in the initial stage. Together with the experiments in Task I and II, it is justifiable to conclude that using images that contain instances of classes from previous stages, even without labels in the incremental learning stage, is essential to ensure successful incremental learning without catastrophic forgetting. These images with relevant instances serve as materials for refreshing and consolidating previously acquired knowledge. %These images with relevant instances serve as materials for ``reviewing'' and strengthen the ``memory'' of past knowledge.

\begin{table}[ht]
        \centering
        \caption{Performance with and without using unlabeled AMOS data in incremental step in Task II.}
        \resizebox{0.9\columnwidth}{!}{%
        \begin{tabular}{c|c|c|c|c}
            \toprule
            & \textbf{Initial Stage} & \multicolumn{2}{c|}{\textbf{Incremental Stage}} &  \textbf{Upper Bound}\\
            & \textbf{Per-structure (1--16)} & \textbf{Per-structure (1--16)} & \textbf{Per-structure (17--19)} & \textbf{Per-structure (17--19)} \\
            \midrule
            w/o unlabeled data & \multirow{2}{*}{\centering 85.2 (12.5)} & $0(0)$ & $86.7(4.2)$ & \multirow{2}{*}{\centering87.1(3.9)}\\
            w/ unlabeled data & & 81.6(15.8) & $86.9(4.1)$ & \\
            \bottomrule
         \end{tabular}}
        \label{tab6}
\end{table}

\section{Limitations}
We acknowledge the following limitations: 1) The Gram-Schmidt algorithm for orthogonalization may restrict its application to tasks where the number of classes is no greater than the feature dimensionality. While increasing the feature dimension can address this, it would require more GPU memory. Another solution is to transform CLIP embeddings or random noises so that the angle between any pair of transformed vectors is sufficiently large, but not necessarily 90 degrees. 2) Our current study only includes one incremental stage, and we have not validated online learning with multiple incremental steps, which is a direction for future work.

\section{Conclusion}
We have developed a scalable approach for training foundational models in automatic medical image segmentation. Our method features a classifier-free neural network, enabling flexibility to accommodate a variable number of classes. We also employ knowledge distillation-based incremental learning, allowing adaptation to non-stationary data streams. Furthermore, we incorporate a model self-disambiguation mechanism, which harnesses partially or sparsely labeled data for training, making it easier to extend capabilities. By addressing the key scalability limitations -- inflexible network architecture, high annotation costs, and dynamic real-world data -- our approach establishes a robust foundation for training universal medical image segmentation models.

%Additionally, we employ knowledge distillation-based incremental learning, enabling adaptation to non-stationary streams of data. Moreover, we integrate a model self-disambiguation mechanism, which leverages more readily available partially or sparsely labeled data for training. By addressing the limiting factors for scalability in network architecture, the high cost of data annotations, and the dynamic nature of real-world data, our approach lays a robust foundation for training universal medical image segmentation models.

%\section*{References}
%\medskip

{
\small
\bibliographystyle{unsrt}
    \bibliography{paper}
}

%%%%%%%%%%%%%%%%%%%%%%%%%%%%%%%%%%%%%%%%%%%%%%%%%%%%%%%%%%%%

\appendix

\section{Appendix / supplemental material}

\subsection{Visual Results}
\begin{figure}[h]
  \centering
  \includegraphics[width=\textwidth]{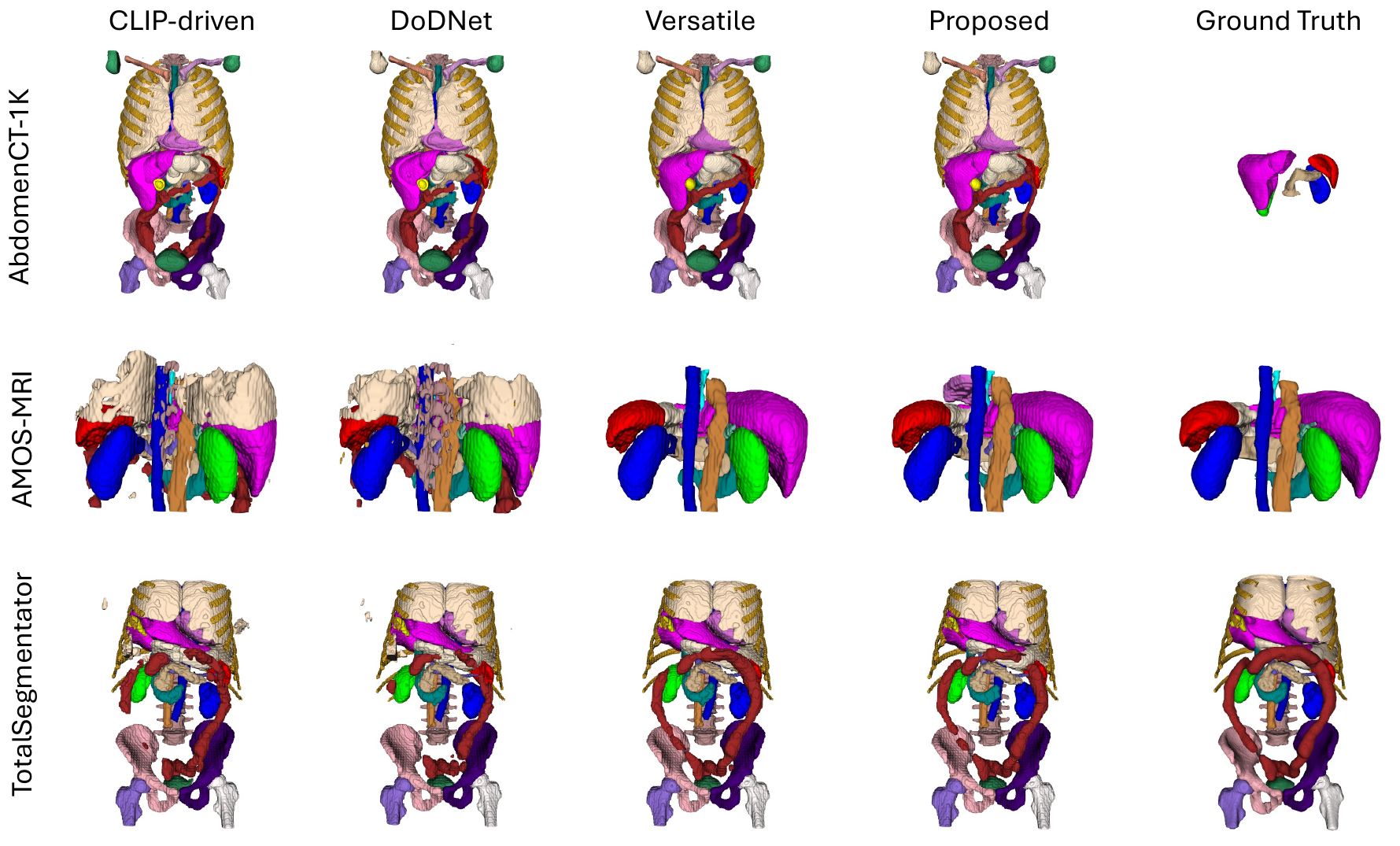}
  \caption{Visual comparison between the ground truth and the predictions generated by DoDNet, CLIP-driven, Versatile model and the proposed method on three subjects from different datasets.}
  \label{fig2}
\end{figure}

We present visual results for the 30-class segmentation task in Fig.~\ref{fig2}, comparing DoDNet, the CLIP-driven model, the Versatile model, and our proposed method. The Versatile model and our proposed method demonstrate comparable segmentation performance, outperforming DoDNet and the CLIP-driven model on most structures across all datasets. Notably, our proposed method successfully segments the heart in MRI images, despite the lack of heart labels in the MRI data used in training, indicating its ability to learn modality-invariant structural information. However, both the Versatile model and our proposed method struggle to segment lung, colon, and vertebrae from MRI images, whereas DoDNet and the CLIP-driven model can partially segment these structures, albeit with imperfect results. This highlights areas for further improvement in our proposed method.

\subsection{More Information about Datasets}
More information about Urogram and LifespanBTS is provided here. We refer the readers to the papers about the seven public datasets for their respective details. 

The Urogram dataset comprises $122$ contrast-enhanced CT images from patients undergoing urinary system examinations. The images have a uniform matrix size of $512 \times 512$, with a variable number of 2D slices ranging from $62$ to $685$. Pixel spacing ranges from $0.607$ to $0.977$ $\text{mm}$, and slice thickness varies from $1.0$ to $3.0$ $\text{mm}$. Urologists annotated the kidney, bladder, and ureters in each image. For this study, only the masks of the two kidneys and the bladder were retained.

The LifespanBTS dataset comprises $75$ T1-weighted brain images with manually corrected brain tissue masks, spanning a broad age range from infants to elderly individuals \cite{chen2023brain}. This dataset is a compilation of images from multiple sources, including the Developing Human Connectome Project (dHCP) \cite{makropoulos2018developing}, Baby Connectome Project (BCP) \cite{howell2019unc}, HCP projects Development (HCP-D) \cite{somerville2018lifespan}, HCP Young Adult (HCP-YA) and HCP Aging (HCP-A) \cite{van2012human}. Specifically, the dataset includes $8$ images from dHCP, $31$ from BCP, $20$ from HCP-D, $10$ from HCP-YA, and $6$ from HCP-A. The images have been manually segmented into GM, WM, and CSF.

Table~\ref{tab7} provides an overview of the annotated structures in each dataset. Notably, none of the datasets are fully annotated, with varying degrees of annotation across the different structures.

\begin{table}[h]
        \centering
        \caption{Annotated anatomical structures in different datasets.}
        \resizebox{\columnwidth}{!}{%
        \begin{tabular}{c|c|c|c|c|c|c|c|c|c|c|c|c|c|c|c|c}
        \toprule
        \textbf{Dataset} & \textbf{Spleen} & \textbf{RK} & \textbf{LK} & \textbf{GB} & \textbf{Eso} & \textbf{Liver} & \textbf{St} & \textbf{Aorta} & \textbf{PC} & \textbf{Pan} & \textbf{RAG} & \textbf{LAG} & \textbf{Duo} & \textbf{Bladder} & \textbf{PU} & \textbf{PSV}\\
        \midrule
        AbdomenCT-1K  & \cmark & \cmark & \cmark & \xmark & \xmark & \cmark & \xmark & \xmark & \xmark & \cmark & \xmark & \xmark & \xmark & \xmark & \xmark & \xmark\\
        AMOS-CT & \cmark & \cmark & \cmark & \cmark & \cmark & \cmark & \cmark & \cmark & \cmark & \cmark & \cmark & \cmark & \cmark & \cmark & \cmark & \xmark\\
        AMOS-MRI & \cmark & \cmark & \cmark & \cmark & \cmark & \cmark & \cmark & \cmark & \cmark & \cmark & \cmark & \cmark & \cmark & \cmark & \cmark & \xmark\\
        BTCV & \cmark & \cmark & \cmark & \cmark & \cmark & \cmark & \cmark & \cmark & \cmark & \cmark & \cmark & \cmark & \xmark & \xmark & \xmark & \cmark\\
        FLARE22 & \cmark & \cmark & \cmark & \cmark & \cmark & \cmark & \cmark & \cmark & \cmark & \cmark & \cmark & \cmark & \cmark & \xmark & \xmark & \xmark\\
        NIH-Pancreas & \xmark & \xmark & \xmark & \xmark & \xmark & \xmark & \xmark & \xmark & \xmark & \cmark & \xmark & \xmark & \xmark & \xmark & \xmark & \xmark\\
        TotalSegmentator & \cmark & \cmark & \cmark & \cmark & \cmark & \cmark & \cmark & \cmark & \cmark & \cmark & \cmark & \cmark & \cmark & \cmark & \xmark & \cmark\\
        Urogram & \xmark & \cmark & \cmark & \xmark & \xmark & \xmark & \xmark & \xmark & \xmark & \xmark & \xmark & \xmark & \xmark & \cmark & \xmark & \xmark\\
        WORD & \cmark & \cmark & \cmark & \cmark & \cmark & \cmark & \cmark & \xmark & \xmark & \cmark & \cmark & \cmark & \cmark & \cmark & \xmark & \xmark\\
        LifespanBTS & \xmark & \xmark & \xmark & \xmark & \xmark & \xmark & \xmark & \xmark & \xmark & \xmark & \xmark & \xmark & \xmark & \xmark & \xmark & \xmark\\
        \bottomrule
        \toprule
        \textbf{LC} & \textbf{RC}& \textbf{Colon} & \textbf{LF} & \textbf{RF} & \textbf{Heart} & \textbf{LH} & \textbf{RH} & \textbf{LHU} & \textbf{RHU} & \textbf{Lung} & \textbf{Rib} & \textbf{Trachea} & \textbf{Vertebrae} & \textbf{GM} & \textbf{WM} & \textbf{CSF}\\
        \midrule
        \xmark & \xmark & \xmark & \xmark & \xmark & \xmark & \xmark & \xmark & \xmark & \xmark & \xmark & \xmark & \xmark & \xmark & \xmark & \xmark & \xmark\\
        \xmark & \xmark & \xmark & \xmark & \xmark & \xmark & \xmark & \xmark & \xmark & \xmark & \xmark & \xmark & \xmark & \xmark & \xmark & \xmark & \xmark\\
        \xmark & \xmark & \xmark & \xmark & \xmark & \xmark & \xmark & \xmark & \xmark & \xmark & \xmark & \xmark & \xmark & \xmark & \xmark & \xmark & \xmark\\
        \xmark & \xmark & \xmark & \xmark & \xmark & \xmark & \xmark & \xmark & \xmark & \xmark & \xmark & \xmark & \xmark & \xmark & \xmark & \xmark & \xmark\\
        \xmark & \xmark & \xmark & \xmark & \xmark & \xmark & \xmark & \xmark & \xmark & \xmark & \xmark & \xmark & \xmark & \xmark & \xmark & \xmark & \xmark\\
        \xmark & \xmark & \xmark & \xmark & \xmark & \xmark & \xmark & \xmark & \xmark & \xmark & \xmark & \xmark & \xmark & \xmark & \xmark & \xmark & \xmark\\
         \cmark & \cmark & \cmark & \cmark & \cmark & \cmark & \cmark & \cmark & \cmark & \cmark & \cmark & \cmark & \cmark & \cmark & \xmark & \xmark & \xmark\\
        \xmark & \xmark & \xmark & \xmark & \xmark & \xmark & \xmark & \xmark & \xmark & \xmark & \xmark & \xmark & \xmark & \xmark & \xmark & \xmark & \xmark\\
        \xmark & \xmark & \xmark & \xmark & \xmark & \xmark & \xmark & \xmark & \xmark & \xmark & \xmark & \xmark & \xmark & \xmark & \xmark & \xmark & \xmark\\
        \xmark & \xmark & \xmark & \xmark & \xmark & \xmark & \xmark & \xmark & \xmark & \xmark & \xmark & \xmark & \xmark & \xmark & \cmark & \cmark & \cmark\\
        \bottomrule
    \end{tabular}}
    \label{tab7}
\end{table}

\subsection{Gram-Schmidt Algorithm}
Let $\mathbf{v}_i$ be the CLIP embedding (after feature selection) for class $i$. We use the following algorithm to obtain the final class prototypes $\{\mathbf{e}_i\}_{i=0}^{N}$, which are unit-length orthogonal vectors. $N$ is the total number of structures of interest.

\begin{align*}
\mathbf{u}_0 &= \mathbf{v}_0 \\
\mathbf{u}_1 &= \mathbf{v}_1 - \text{proj}_{\mathbf{u}_0}(\mathbf{v}_1) \\
\mathbf{u}_2 &= \mathbf{v}_2 - \text{proj}_{\mathbf{u}_0}(\mathbf{v}_2) - \text{proj}_{\mathbf{u}_1}(\mathbf{v}_2) \\
& \vdots \\
\mathbf{u}_N &= \mathbf{v}_N - \sum_{i=0}^{N-1} \text{proj}_{\mathbf{u}_{i}}(\mathbf{v}_N)
\end{align*}

where $\text{proj}_{\mathbf{u}}(\mathbf{v}) =  \frac{\langle \mathbf{v}, \mathbf{u} \rangle}{\langle \mathbf{u}, \mathbf{u} \rangle} \mathbf{u}$, $\langle \mathbf{u}, \mathbf{v} \rangle$ represents the inner product of $\mathbf{u}$ and $\mathbf{v}$. $\mathbf{e}_i = \frac{\mathbf{u}_i}{\| \mathbf{u}_i \|}$, and $\|\cdot\|$ denotes the norm.

\subsection{Ablation Study}
\noindent \textbf{Impact of $\tau$} We conducted experiments with three different values for $\tau$ in both 16-class and 30-class segmentation tasks, using partially labeled data from eight datasets: AbdomenCT-1K, AMOS (excluding unlabeled images), BTCV, FLARE22, NIH pancreas, TotalSegmentator, WORD, and Urogram. As shown in Table~\ref{tab8}, he segmentation performance remains robust and insensitive to $\tau$ when its value falls within the range of [0.05, 0.12]. In this paper, we adopt a default value of $0.12$ for $\tau$ in our experiments, as it yields slightly superior performance compared to other values.

\begin{table}[h]
        \centering
        \caption{Method performance comparison on partially labeled data with different $\tau$.}
        \resizebox{0.66\columnwidth}{!}{%
        \begin{tabular}{c|c|c|c|c}
            \toprule
            \multirow{2}{*}{\centering \textbf{$\tau$}}  & \multicolumn{2}{c|}{\textbf{16-class Task}} & \multicolumn{2}{c}{\textbf{30-class Task}} \\
            & \textbf{Per-subject} & \textbf{Per-structure} & \textbf{Per-subject} & \textbf{Per-structure} \\
            \midrule
            0.05 & $88.4 (7.8)$ & $85.1 (12.6)$ & $89.3 (5.7)$ & $88.6 (9.6)$ \\
            0.07 & $88.5 (7.7)$ & $85.2 (12.6)$ & $89.2 (5.8)$ & $88.4 (10.0)$ \\
            0.12 & $88.6 (7.1)$ & $85.4 (12.4)$ & $89.5 (5.6)$ & $88.5 (10.0)$ \\
            \bottomrule
         \end{tabular}}
        \label{tab8}
\end{table}

\noindent \textbf{Random Noises vs. CLIP Embeddings} Our experiments, with $\tau$ set to $0.12$ in both 16-class and 30-class tasks, revealed that using random noises and CLIP embeddings to generate class prototypes yields comparable results, with no substantial differences observed between the two approaches, as shown in Table~\ref{tab9}.

\begin{table}[h]
        \centering
        \caption{Method performance comparison between random noises and CLIP embeddings ($\tau$ = 0.12).}
        \resizebox{0.8\columnwidth}{!}{%
        \begin{tabular}{c|c|c|c|c}
            \toprule
            \multirow{2}{*}{\centering \textbf{Approach}}  & \multicolumn{2}{c|}{\textbf{16-class Task}} & \multicolumn{2}{c}{\textbf{30-class Task}} \\
            & \textbf{Per-subject} & \textbf{Per-structure} & \textbf{Per-subject} & \textbf{Per-structure} \\
            \midrule
            Random noises & $88.4 (7.6)$ & $85.5 (12.4)$ & $89.3 (5.9)$ & $88.2 (10.2)$ \\
            CLIP embeddings & $88.6 (7.1)$ & $85.4 (12.4)$ & $89.5 (5.6)$ & $88.5 (10.0)$ \\
            \bottomrule
         \end{tabular}}
        \label{tab9}
\end{table}

\begin{figure}[h]
  \centering
  \includegraphics[width=0.8\textwidth]{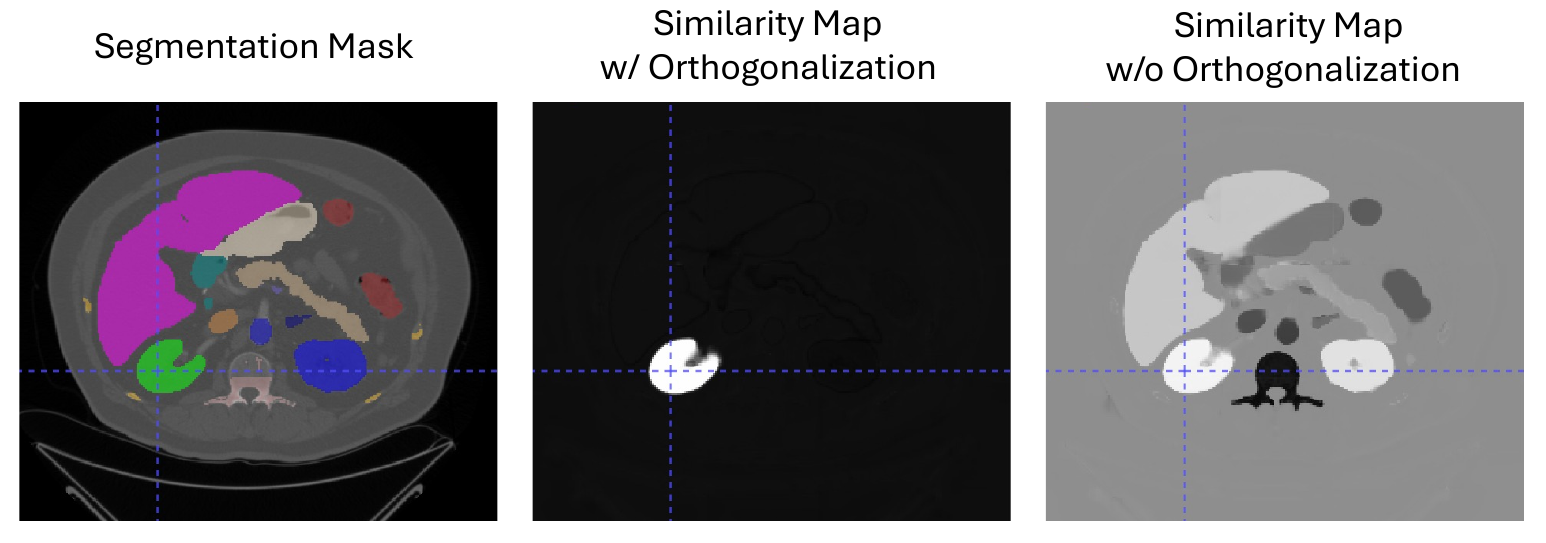}
  \caption{Similarity map calculated as the cosine similarity between the model's predicted feature representations and the right kidney's class prototype.}
  \label{fig3}
\end{figure}

\noindent \textbf{With vs. Without Orthogonalization} As shown in Table~\ref{tab10}, we found that the model without using Gram-Schmidt algorithm can only get comparable performance to model with Gram-Schmidt algorithm being used when the $\tau$ is sufficiently small. Orthogonalization enlarges the range of suitable values for $\tau$, facilitating model training. In addition, as shown in Fig.~\ref{fig3}, the similarity maps, which were generated by calculating the cosine similarity between the model's predicted feature representations and the right kidney's class prototype, with and without orthogonalization, reveal that orthogonalization leads to more discriminative features. This is evident in the similarity maps, where the right kidney region is highlighted in bright colors (with values close to 1), indicating strong similarity, while the other regions are shown in dark colors (with values close to 0), indicating weak similarity. In contrast, the cosine similarity between the feature representations of the right kidney and its corresponding class prototype is approximately $0.2$, significantly deviating from the ideal value of $1$. This disparity suggests that the learned features do not closely align with the predefined class prototypes.

\begin{table}[h]
        \centering
        \caption{Method performance comparison between with and without orthogonalization on partially labeled data.}
        \resizebox{0.8\columnwidth}{!}{%
        \begin{tabular}{c|c|c|c|c|c}
            \toprule
            & \multirow{2}{*}{\textbf{$\tau$}} & \multicolumn{2}{c|}{\textbf{16-class Task}} & \multicolumn{2}{c}{\textbf{30-class Task}} \\
            & & \textbf{Per-subject} & \textbf{Per-structure} & \textbf{Per-subject} & \textbf{Per-structure} \\
            \midrule
            \multirow{3}{*}{w/o orthogonalization} & 0.03 & $88.2(7.5)$ & $85.2(12.5)$ & $87.9(6.4)$ & $86.9(11.0)$ \\
                                                                        & 0.05 & $83.6(10.7)$ & $79.0(14.8)$ & $60.5(8.8)$ & $56.2(13.3)$ \\
                                                                        & 0.07 & $76.1(15.0)$ & $67.3(15.6)$ & $69.4(15.3)$ & $52.7(13.3)$ \\
                                                                        & 0.12 & $74.0(16.0)$ & $64.6(16.0)$ & $58.4(8.4)$ & $51.8(12.7)$ \\
            \midrule
            \multirow{3}{*}{w/ orthogonalization} & 0.03 & $-$ & $-$ & $-$ & $-$ \\
                                                                      & 0.05 & $88.4 (7.8)$ & $85.1 (12.6)$ & $89.3 (5.7)$ & $88.6 (9.6)$ \\
                                                                      & 0.07 & $88.5 (7.7)$ & $85.2 (12.6)$ & $89.2 (5.8)$ & $88.4 (10.0)$ \\
                                                                      & 0.12 & $88.6 (7.1)$ & $85.4 (12.4)$ & $89.5 (5.6)$ & $88.5 (10.0)$ \\
            \bottomrule
         \end{tabular}}
        \label{tab10}
\end{table}

\noindent \textbf{Fixed vs. Learnable $\tau$} As shown in Table~\ref{tab11}, we found that when the $\tau$ is learnable, the segmentation performance is sensitive to the initial value. However, when it is properly initialized, we see no significant difference in segmentation performance between learnable and fixed $\tau$. Similar to the findings in the similarity maps in experiments without orthogonalization, the cosine similarity between the feature representations and their corresponding class prototypes falls significantly short of the optimal value of $1$. This deviation implies a lack of strong alignment between the learned features and predefined class prototypes.

\begin{table}[h!]
        \centering
        \caption{Method performance comparison between fixed and learnable $\tau$.}
        \resizebox{0.5\columnwidth}{!}{%
        \begin{tabular}{c|c|c|c}
            \toprule
            & \multirow{2}{*}{\textbf{Initial Value}} & \multicolumn{2}{c}{\textbf{30-class Task}} \\
            & & \textbf{Per-subject} & \textbf{Per-structure} \\
            \midrule
            \multirow{3}{*}{Learnable} & 0.05 & $89.2(5.7)$ & $88.4(10.0)$ \\
                                                      & 0.07 & $89.3(5.6)$ & $88.5(9.7)$ \\
                                                      & 0.12 & $86.4(7.6)$ & $83.2(11.4)$ \\
            \midrule
            Fixed & - & $89.5 (5.6)$ & $88.5 (10.0)$ \\
            \bottomrule
         \end{tabular}}
        \label{tab11}
\end{table}

\subsection{Adaptation and Training of Compared Methods for Incremental Learning}
The three methods compared with our proposed approach -- DoDNet, the CLIP-driven model, and the Versatile model -- were not originally designed for incremental learning. To demonstrate the superiority of our proposed method in incremental learning, we adapted these methods for incremental learning and used them for comparative analysis in this study. Specific training details for adapting them to incremental learning are as follows:

\noindent \textbf{Incremental Learning Procedure for DoDNet} 1) Initialize a teacher model and train it on the initial stage data, using a one-hot vector of length equal to the total number of classes involved in both the initial and incremental stages plus one. This ensures the teacher model has the same architecture as the student model, which is trained in the subsequent stage. 2) Create a student model and initialize it with the weights of the teacher model. 3) Select a training example from the incremental stage data and a task identifier from the newly added label set $\Phi_{N+1}^{N+M}$ for the student model and calculate the binary cross-entropy loss and dice loss based on the network output and the ground truth. 4) Feed the same data and a task identifier from the initial stage label set $\Phi_{N}$ into both the frozen teacher model and the student model to obtain their respective predictions, which are used for computing the knowledge distillation loss. Note that the knowledge distillation loss involves all voxels and differs from Equations 7 and 8, as the model performs conditional segmentation with a single-channel output. 5) Train the student model using a weighted combination of the binary cross-entropy loss, dice loss, and knowledge distillation loss.

\noindent \textbf{Incremental Learning Procedure for the CLIP-driven Model} The training procedure for the CLIP-driven model is largely identical to that of DoDNet, with one key distinction: the first step involves initializing and training a teacher model without additional considerations for the architecture, as the CLIP-driven model's architecture does not depend on the number of classes.

\noindent \textbf{Incremental Learning Procedure for the Versatile Model} 1) Initialize a teacher model and train it on the initial stage data. 2) Create a student model and initialize all its layers, except for the last convolutional layer, with the weights of the corresponding layer in the teacher model. For the last convolutional layer, initialize the first $N+1$ convolutional kernels and biases with the weights from the teacher model's last convolutional layer. Randomly initialize all other kernels in this layer. 3) Select a training example from the incremental stage data and feed it into both the frozen teacher model and the student model to obtain their respective predictions. 4) Calculate the ambiguity-aware cross-entropy loss, and the ambiguity-aware dice loss based on the student model's output and the ground truth. Additionally, compute the entropy minimization and volume minimization terms using the student model's output. 5) Calculate the knowledge distillation loss using pre-softmax logits. This loss measures the $L_{1}$ distance between the logits from the teacher and student models. In our experiments, we found that balancing the knowledge distillation loss among classes was more effective. To achieve this, use pseudo-labels from the teacher model to create a binary mask for each class in $\Phi_{N}$. Then, calculate the mean distance between logits from the two models within each class region and balance the contribution of different classes by taking the mean of these mean losses.

Note that the incremental learning procedure of the proposed model differs from that of the Versatile models in two key aspects. Firstly, both the teacher and student models share identical architecture and have the same number of parameters. Consequently, the student model can be initialized using all weights from the teacher model, eliminating the need to train any parameters from scratch. Secondly, we observed that using the knowledge distillation loss as defined in Equations 7 and 8, which simply computes the average Kullback–Leibler divergence between the teacher model's probability distributions and those of the student models (after background remodeling) for the incremental stage training data, is highly effective.

%%%%%%%%%%%%%%%%%%%%%%%%%%%%%%%%%%%%%%%%%%%%%%%%%%%%%%%%%%%%

\end{document}